\documentclass[letterpaper]{article} 
\usepackage{aaai2027} 
\usepackage[hyphens]{url}  
\usepackage{graphicx} 
\usepackage{natbib}  
\usepackage{caption} 
\usepackage{algorithm}
\usepackage{algorithmic}
\usepackage{pgfplots}
\usepgfplotslibrary{groupplots}
\pgfplotsset{compat=1.17}
\usetikzlibrary{patterns}
\usepackage{amsfonts}

\usepackage{newfloat}
\usepackage{listings}
\DeclareCaptionStyle{ruled}{labelfont=normalfont,labelsep=colon,strut=off} 
\floatstyle{ruled}
\newfloat{listing}{tb}{lst}{}
\floatname{listing}{Listing}
\usepackage{tikz}
\usetikzlibrary{calc,positioning,arrows.meta,backgrounds,fit,shadows,decorations.pathreplacing}
\usepackage{pgfplots}
\pgfplotsset{compat=1.18}
\usepackage{nicematrix} 
\usepackage{pgfplotstable}
\usetikzlibrary{positioning, fit, arrows.meta, shapes.multipart}
\pgfplotsset{compat=1.18}

\definecolor{tydracolor}{HTML}{0F766E}
\definecolor{tabpfncolor}{HTML}{D9A05B}
\definecolor{hydra22Mcolor}{HTML}{7A9BD1}
\definecolor{hydra160Mcolor}{HTML}{1A4B9B}
\definecolor{lightgreen}{RGB}{198, 230, 200}
\usepackage{tikz}          
\usepackage[table]{xcolor}
\usepackage{pgfmath}
\usepackage{etoolbox}

\colorlet{high}{blue!70!black}
\colorlet{low}{white}
\def\opacity{30}

\colorlet{greenlow}{green!10}
\colorlet{greenhigh}{green!70!black}

\newcommand{\applygradientgreen}[5]{%
    \pgfmathparse{int(round(100*(#1-#4)/(#5-#4)))}%
    \xdef\tempcolor{\pgfmathresult}%
    \cellcolor{greenhigh!\tempcolor!greenlow!\opacity}%
    \ifnum\tempcolor>80\color{white}\else\color{black}\fi%
    \ifstrequal{#3}{bold}{\textbf{#1}}{#1}%
    \if\relax\detokenize{#2}\relax\else\;\scriptsize #2\fi
}
\newcommand{\applygradient}[5]{%
    \pgfmathparse{int(round(100*(#1-#4)/(#5-#4)))}%
    \xdef\tempcolor{\pgfmathresult}%
    \cellcolor{high!\tempcolor!low!\opacity}%
    \ifnum\tempcolor>80\color{white}\else\color{black}\fi%
    \ifstrequal{#3}{bold}{\textbf{#1}}{#1}%
    \if\relax\detokenize{#2}\relax\else\;\scriptsize #2\fi
}

\title{Tydra: An Efficient Hybrid Model for Tabular Data}

\author {
    Mieszko Komisarczyk\textsuperscript{\rm 1}\equalcontrib, 
    Saurabh Mathur\textsuperscript{\rm 1}\equalcontrib, 
    Maurice Kraus\textsuperscript{\rm 1}, 
    Sriraam Natarajan\textsuperscript{\rm 2}, 
    Kristian~Kersting\textsuperscript{\rm 1,3,4}
}
\affiliations {
    \textsuperscript{\rm 1} Department of Computer Science, Technical University of Darmstadt, Germany\\
    \textsuperscript{\rm 2}Department of Computer Science, The University of Texas at Dallas, USA\\
    \textsuperscript{\rm 3} Hessian Center for Artificial Intelligence (hessian.ai), Darmstadt, Germany \\
    \textsuperscript{\rm 4}German Research Center for AI (DFKI)
}

\nocopyright

\begin{document}

\maketitle
\begin{abstract}

Transformer-based tabular foundation models such as TabPFN achieve strong predictive performance but incur quadratic computational cost with context length. On the other hand, subquadratic SSM-based alternatives such as Hydra trade away accuracy for efficiency. To balance both, we introduce Tydra, a hybrid Transformer–State Space Model (SSM) architecture for tabular in-context learning that interleaves attention and SSM layers.
Across 30 OpenML datasets, Tydra reduces inference time by 30\% relative to TabPFN while retaining much of its predictive performance. Tydra also outperforms an approximately ten-times-larger Hydra model while providing faster inference. The results indicate that hybrid architectures are a promising direction for tabular foundation models.

\end{abstract}

\section{Introduction}
Tabular foundation models such as TabPFN have achieved strong predictive performance on several tabular prediction tasks without task-specific training~\cite{hollmann2025accurate}. However, TabPFN's Transformer backbone scales quadratically with context length, making inference prohibitively expensive on large-scale or long-context tabular tasks. This cost is especially restrictive for institutions that cannot use server-hosted inference at all. Hospitals and other regulated institutions are often barred from sending patient records to an external server for data protection reasons and must instead run inference locally on hardware with limited compute resources. While the Hydra architecture reduces this cost with a subquadratic State Space Model (SSM) architecture, its predictive performance falls short of TabPFN~\cite{hwang2024hydra,koch2025state}. This gap persists even as Hydra is scaled up; shrinking TabPFN to match Hydra's efficiency degrades its accuracy. Neither pure architecture achieves both high accuracy and low inference cost.
\begin{figure}[t]
    \centering
    \includegraphics[width=\linewidth]{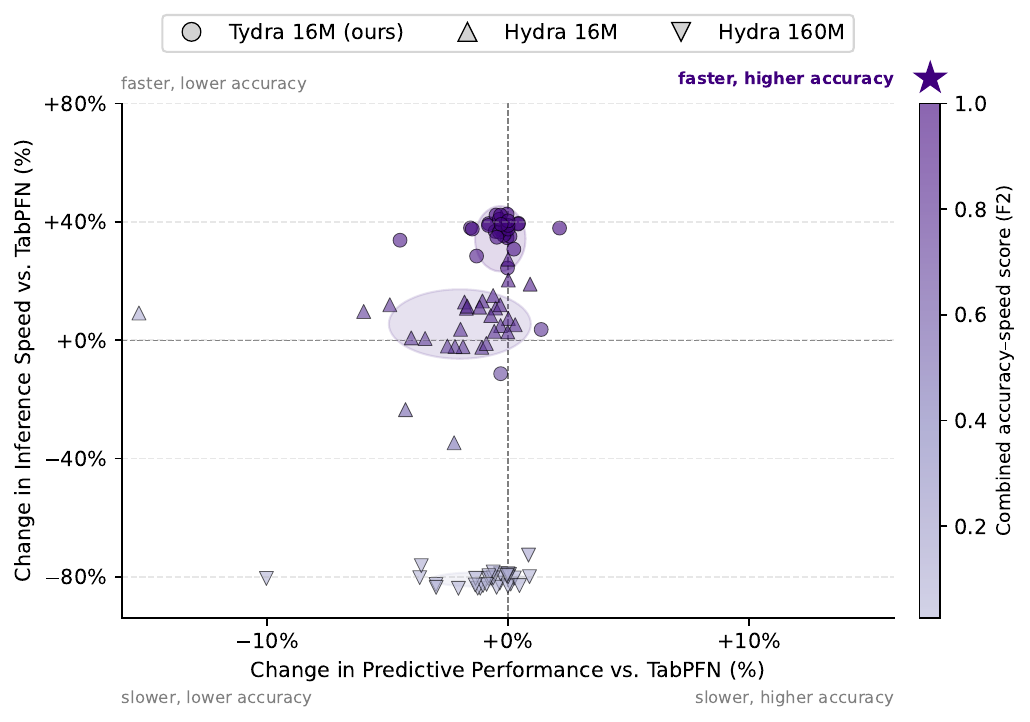}
   \caption{
    \textbf{Tydra is fast on tabular data without sacrificing predictive performance.} Inference speed and predictive performance of Tydra compared to TabPFN and Hydra on OpenML datasets. Color encodes a combined accuracy–speed score (F2, weighted toward speed). Shaded ellipses span one standard deviation around each model's mean. The star marks the ideal corner: fastest and most accurate. Hydra 16M is fast but trails substantially in predictive performance, while TabPFN and Hydra 160M are accurate but slow; Tydra achieves predictive performance close to TabPFN and Hydra 160M at over $2\times$ their inference speed.}\label{fig:quadrants}
    \label{fig:placeholder}
    \vspace{-.1cm}
\end{figure}

Hybrid architectures that interleave attention and SSM layers have recently improved this accuracy–efficiency trade-off for language models \citep{merrill2026olmo}, though such architectures remain unexplored for tabular foundation models. Tabular data are permutation-invariant over rows and columns, unlike the fixed-order sequences of language for these hybrid architectures were designed. To this end, we introduce Tydra, which interleaves TabPFN's attention layers with Hydra's SSM layers. We show that this hybridization {\bf matches TabPFN's accuracy at substantially lower inference cost on small-to-medium-scale data}, making strong tabular in-context learning practical in settings where server-hosted inference is not an option. 

To summarize, we make the following contributions:
 \begin{description}
   \item{\bf (1)} We introduce Tydra, the first hybrid Transformer–SSM architecture for tabular in-context learning.
   \item{\bf (2)} We show that Tydra matches TabPFN's accuracy at up to 30\% lower inference time on 30 OpenML datasets, while substantially outperforming Hydra.  
\item{\bf (3)} We provide an extensive study on the architecture family of Tydras showing the advantages of different combinations and ratios of Hydra and TabPFN layers. 
\end{description}
We proceed as follows. We start off by discussing related work. We then present the Tydra architecture. Before concluding, we present empirical results.



\section{Related Work}
Tydra is related to several lines of works, namely, TabPFN, state space models, and hybrid models.

\subsection{TabPFN}
TabPFN (Tabular Prior-Data Fitted Network) is a tabular foundation model that performs tabular classification via in-context learning. Given an entire training set and test queries, it makes predictions via a single forward pass. Its architecture is based on the Transformer, with self-attention adapted for permutation invariance across rows and columns. This adapted Transformer is meta-trained offline on millions of synthetic classification tasks, yielding strong predictive accuracy across a wide range of tabular datasets without any dataset-specific training. However, since self-attention scales quadratically with the number of rows in a tabular dataset, inference can be computationally expensive.

Since the original release, several extensions have targeted TabPFN's scope and scaling. TabPFN~v2~\citep{hollmann2025accurate} broadened the original classification-only model~\citep{hollmann2022tabpfn} to regression and increased model capacity, while later revisions, TabPFN~v2.5 and v3~\citep{grinsztajn2026tabpfn} roughly doubled transformer depth and raised the supported class count. These successive versions improve accuracy and task coverage, but retain the same attention mechanism and therefore inherit its quadratic cost in the number of rows. Recognizing this bottleneck, a separate line of work has proposed \emph{post-hoc} strategies to extend TabPFN's practical reach without retraining or architectural change: subsampling and divide-and-conquer schemes that partition large datasets into TabPFN-sized chunks and aggregate predictions~\citep{ye2026closer}, and domain-specific adaptations such as TabPFN-TS for time-series forecasting~\citep{hoo2025tables}. These efforts confirm that the scaling limitation is widely recognized, but they work around the transformer's row-scaling behavior rather than removing it. In contrast, our approach addresses the bottleneck at the architectural level.

\subsection{State Space Models} State Space Models (SSMs) have emerged as an effective approach to overcoming the limitations of both RNNs and Transformers. Derived from continuous-time dynamical systems, they achieve near-linear complexity in sequence length \citep{somvanshi2025advancing}. Mamba \citep{gu2023mamba} is a prominent example, introducing a selective state space mechanism that adaptively filters information across the sequence; this selectivity allows Mamba to achieve linear-time inference while matching or exceeding Transformer performance across several domains \citep{gu2023mamba}. Hydra \citep{hwang2024hydra} extends Mamba to non-causal settings via quasiseparable matrix mixers, enabling bidirectional context aggregation while retaining Mamba's efficiency --- a property particularly relevant for tabular data, where rows have no natural sequence order. Hydra has previously been adapted to the tabular setting \citep{koch2025state}. 
While TabPFN outperforms Hydra in both inference speed and predictive accuracy on standard small-to-medium-scale datasets, Hydra becomes the better choice at larger dataset sizes, where TabPFN's quadratic cost becomes prohibitive and Hydra's subquadratic scaling allows it to remain practical. This suggests that SSMs offer a genuine efficiency advantage over attention-based tabular foundation models but at the cost of predictive performance, especially in standard small-to-medium scale datasets.



\subsection{Hybrid models} Hybrid architectures that combine transformer layers with more efficient, subquadratic layers have emerged as an effective solution for the efficiency--accuracy tradeoff. Hybrid language models have achieved promising results across common reasoning, math, and coding tasks \citep{merrill2026olmo, li2026transmamba}. Several hybridization strategies have been proposed, including interleaving transformer and state-space blocks \citep{lieber2024jamba}, reusing a small number of attention blocks across an otherwise recurrent backbone \citep{glorioso2024zamba}, and computing recurrent and attention branches in parallel over the same input \citep{dong2025hymba}. However, hybridization has so far been explored almost exclusively for language modeling, with only limited extensions to other modalities \citep{zhu2024samba}, and, to our knowledge, no prior work has introduced a hybrid architecture for tabular foundation models. \\

Tydra precisely addresses this gap, adopting the interleaved strategy to combine TabPFN's attention layers with Hydra's state-space layers.

\section{The Tydra Family of Architectures}

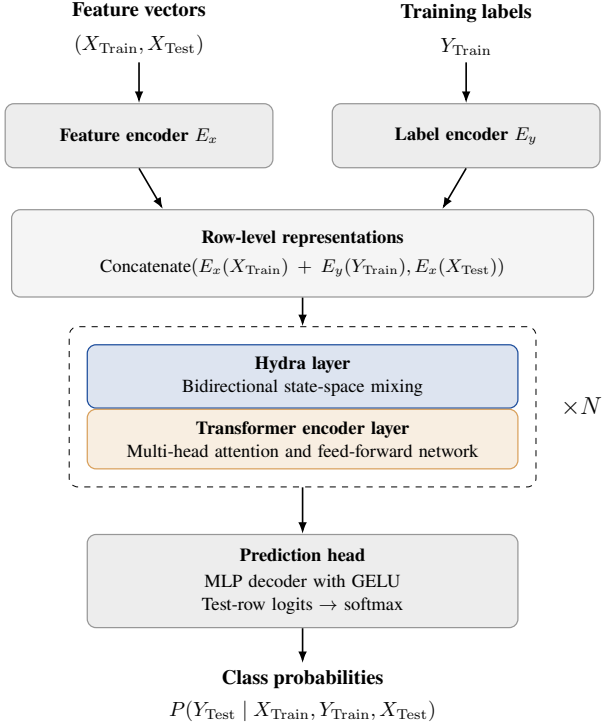
\begin{figure}[t]
\centering
\definecolor{hydrablock}{HTML}{1A4B9B}
\definecolor{transblock}{HTML}{D9A05B}
\definecolor{ioblock}{HTML}{6B6B6B}

\resizebox{0.95\linewidth}{!}{%
\begin{tikzpicture}[
    font=\small,
    >=Latex,
    node distance=4mm,
    input/.style={
        align=center,
        font=\bfseries
    },
    encoder/.style={
        draw=ioblock!70,
        rounded corners=1.5mm,
        fill=ioblock!12,
        minimum height=11mm,
        text width=42mm,
        inner sep=2mm,
        align=center,
        line width=0.6pt
    },
    representation/.style={
        draw=ioblock!70,
        rounded corners=1.5mm,
        fill=ioblock!7,
        minimum height=15mm,
        text width=96mm,
        inner sep=2mm,
        align=center,
        line width=0.6pt
    },
    hydra/.style={
        draw=hydrablock,
        rounded corners=1.5mm,
        fill=hydrablock!15,
        minimum height=10mm,
        text width=70mm,
        inner sep=2mm,
        align=center,
        line width=0.6pt
    },
    transformer/.style={
        draw=transblock,
        rounded corners=1.5mm,
        fill=transblock!18,
        minimum height=10mm,
        text width=70mm,
        inner sep=2mm,
        align=center,
        line width=0.6pt
    },
    prediction/.style={
        draw=ioblock!70,
        rounded corners=1.5mm,
        fill=ioblock!12,
        minimum height=16mm,
        text width=70mm,
        inner sep=2mm,
        align=center,
        line width=0.6pt
    },
    output/.style={
        align=center,
        font=\bfseries
    },
    arrow/.style={
        -{Latex[length=2mm]},
        line width=0.8pt
    }
]


\node[encoder] (xencoder) {
    \textbf{Feature encoder} $E_x$\\[1.5mm]
};

\node[
    encoder,
    right=10mm of xencoder
] (yencoder) {
    \textbf{Label encoder} $E_y$\\[1.5mm]
};


\node[
    input,
    above=7mm of xencoder
] (features) {
    \textbf{Feature vectors}\\[2mm]
    $\left(
        X_{\mathrm{Train}},
        X_{\mathrm{Test}}
    \right)$
};

\node[
    input,
    above=7mm of yencoder
] (labels) {
    \textbf{Training labels}\\[2mm]
    $Y_{\mathrm{Train}}$
};

\draw[arrow] (features) -- (xencoder);
\draw[arrow] (labels) -- (yencoder);


\coordinate (encoder-midpoint) at
    ($(xencoder.south)!0.5!(yencoder.south)$);

\node[
    representation,
    below=7mm of encoder-midpoint
] (representations) {
    \textbf{Row-level representations}\\[1.5mm]

    $\displaystyle
        \text{Concatenate}(
        E_x(X_{\mathrm{Train}})
        +
        E_y(Y_{\mathrm{Train}}),
        E_x(X_{\mathrm{Test}}))
    $
    
};

\draw[arrow]
    (xencoder.south) --
    ([xshift=-24mm]representations.north);

\draw[arrow]
    (yencoder.south) --
    ([xshift=24mm]representations.north);


\node[
    hydra,
    below=8mm of representations
] (hydra-layer) {
    \textbf{Hydra layer}\\[0.5mm]
    {\footnotesize Bidirectional state-space mixing}
};

\node[
    transformer,
    below=0mm of hydra-layer
] (transformer-layer) {
    \textbf{Transformer encoder layer}\\[0.5mm]
    {\footnotesize
        Multi-head attention and feed-forward network
    }
};

\node[
    draw,
    dashed,
    rounded corners=1.5mm,
    inner sep=3mm,
    fit=(hydra-layer)(transformer-layer)
] (hybrid-pair) {};

\node[
    right=3mm of hybrid-pair,
    font=\Large
] (repeat) {
    $\times N$
};

\draw[arrow]
    (representations.south) --
    (hybrid-pair.north);


\node[
    prediction,
    below=8mm of hybrid-pair
] (prediction-head) {
    \textbf{Prediction head}\\[1mm]
    {\footnotesize
        MLP decoder with GELU
    }\\[0.5mm]
    {\footnotesize
        Test-row logits
        $\rightarrow$
        softmax
    }
};

\draw[arrow]
    (hybrid-pair.south) --
    (prediction-head.north);


\node[
    output,
    below=6mm of prediction-head
] (probabilities) {
    \textbf{Class probabilities}\\[1.5mm]
    $\displaystyle
        P\!\left(
            Y_{\mathrm{Test}}
            \mid
            X_{\mathrm{Train}},
            Y_{\mathrm{Train}},
            X_{\mathrm{Test}}
        \right)
    $
};

\draw[arrow]
    (prediction-head.south) --
    (probabilities.north);

\end{tikzpicture}%
}
\caption{\textbf{Tydra's Hybrid Architecture.}
Feature vectors are projected into row-level embeddings, with label
embeddings added for training rows. The resulting sequence passes through
four Hydra--Transformer encoder pairs (eight layers total), each ordered
as Hydra followed by attention. An MLP decoder produces class logits for
the test rows, which are converted into class probabilities by softmax.}

\label{fig:hybrid_architecture}
\end{figure}

\usetikzlibrary{decorations.pathreplacing}

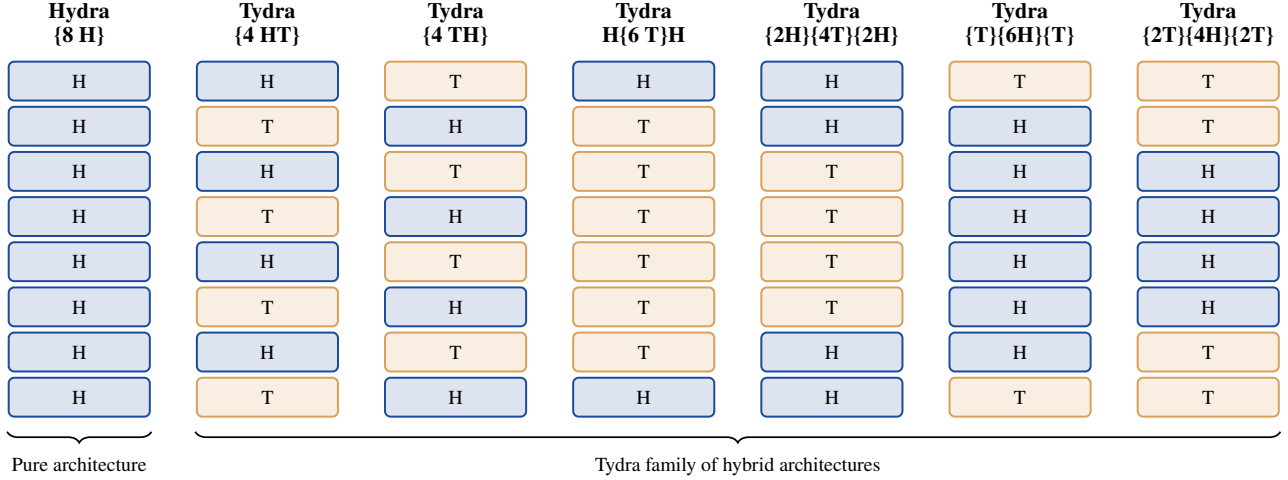
\begin{figure*}[t]
\centering

\definecolor{hydrablock}{HTML}{1A4B9B}
\definecolor{transblock}{HTML}{D9A05B}

\resizebox{0.95\textwidth}{!}{%
\begin{tikzpicture}[
    layer/.style={
        draw=black,
        line width=0.9pt,
        rounded corners=2.5pt,
        minimum width=2.25cm,
        minimum height=0.62cm,
        inner sep=0pt,
        font=\small
    },
    title/.style={
        align=center,
        font=\bfseries
    },
    legendtext/.style={
        anchor=west,
        font=\small
    }
]


\newcommand{\HBlock}[2]{%
    \node[
        layer,
        draw=hydrablock,
        fill=hydrablock!15,
        text=black
    ] at (#1,#2) {H};
}

\newcommand{\TBlock}[2]{%
    \node[
        layer,
        draw=transblock,
        fill=transblock!18,
        text=black
    ] at (#1,#2) {T};
}


\node[title] at (0,5.95) {
    Hydra\\
    \{8 H\}
};

\HBlock{0}{0.00}
\HBlock{0}{0.72}
\HBlock{0}{1.44}
\HBlock{0}{2.16}
\HBlock{0}{2.88}
\HBlock{0}{3.60}
\HBlock{0}{4.32}
\HBlock{0}{5.04}


\node[title] at (3,5.95) {
    Tydra\\
    \{4 HT\}
};

\TBlock{3}{0.00}
\HBlock{3}{0.72}
\TBlock{3}{1.44}
\HBlock{3}{2.16}
\TBlock{3}{2.88}
\HBlock{3}{3.60}
\TBlock{3}{4.32}
\HBlock{3}{5.04}


\node[title] at (6,5.95) {
    Tydra\\
    \{4 TH\}
};

\HBlock{6}{0.00}
\TBlock{6}{0.72}
\HBlock{6}{1.44}
\TBlock{6}{2.16}
\HBlock{6}{2.88}
\TBlock{6}{3.60}
\HBlock{6}{4.32}
\TBlock{6}{5.04}


\node[title] at (9,5.95) {
    Tydra\\
    H\{6 T\}H
};

\HBlock{9}{0.00}
\TBlock{9}{0.72}
\TBlock{9}{1.44}
\TBlock{9}{2.16}
\TBlock{9}{2.88}
\TBlock{9}{3.60}
\TBlock{9}{4.32}
\HBlock{9}{5.04}


\node[title] at (12,5.95) {
    Tydra\\
    \{2H\}\{4T\}\{2H\}
};

\HBlock{12}{0.00}
\HBlock{12}{0.72}
\TBlock{12}{1.44}
\TBlock{12}{2.16}
\TBlock{12}{2.88}
\TBlock{12}{3.60}
\HBlock{12}{4.32}
\HBlock{12}{5.04}


\node[title] at (15,5.95) {
    Tydra\\
    \{T\}\{6H\}\{T\}
};

\TBlock{15}{0.00}
\HBlock{15}{0.72}
\HBlock{15}{1.44}
\HBlock{15}{2.16}
\HBlock{15}{2.88}
\HBlock{15}{3.60}
\HBlock{15}{4.32}
\TBlock{15}{5.04}


\node[title] at (18,5.95) {
    Tydra\\
    \{2T\}\{4H\}\{2T\}
};

\TBlock{18}{0.00}
\TBlock{18}{0.72}
\HBlock{18}{1.44}
\HBlock{18}{2.16}
\HBlock{18}{2.88}
\HBlock{18}{3.60}
\TBlock{18}{4.32}
\TBlock{18}{5.04}


\draw[
    decorate,
    decoration={
        brace,
        mirror,
        amplitude=5pt
    },
    line width=0.8pt
]
(-1.15,-0.55) -- (1.15,-0.55)
node[
    midway,
    below=8pt,
    font=\small
] {
    Pure architecture
};

\draw[
    decorate,
    decoration={
        brace,
        mirror,
        amplitude=5pt
    },
    line width=0.8pt
]
(1.85,-0.55) -- (19.15,-0.55)
node[
    midway,
    below=8pt,
    font=\small
] {
    Tydra family of hybrid architectures
};

\end{tikzpicture}%
}

\caption{
\textbf{The Tydra Family of Architectures.} Architectures of the evaluated Hydra and Tydra models.
Blocks are ordered from the input layer at the bottom to the output
layer at the top. \colorbox{transblock!18}{T} denotes a Transformer attention layer,
and \colorbox{hydrablock!18}{H} denotes a Hydra layer.
}
\label{fig:tydra_architectures}

\end{figure*}

We now present \textit{Tydra}, a family of hybrid architectures for tabular in-context learning that interleaves attention layers with Hydra's state-space (SSM) layers. This section describes Tydra's architecture, model initialization, and training.


\subsection{Hybrid Architecture} Figure~\ref{fig:hybrid_architecture} illustrates Tydra\footnote{We provide the code and implementation details in the Appendix.}, our hybrid
prior-fitted architecture. Without losing of generality we refer to the fully interleave Tydra \{4 HT\} model as Tydra. Other variants of Tydra can be represented by modifying the backbone part.

Given training features
$X_{\mathrm{train}}$, training labels $Y_{\mathrm{train}}$, and test
features $X_{\mathrm{test}}$, Tydra estimates
\[
p\!\left(Y_{\mathrm{test}}
\mid X_{\mathrm{train}},Y_{\mathrm{train}},X_{\mathrm{test}}\right).
\]
Tydra combines the efficient bidirectional sequence mixing of Hydra
with the content-dependent interactions provided by self-attention.

Each table row is represented by one token. Features and labels are
embedded separately using linear encoders,
\begin{equation*}
E_x:\mathbb{R}^{n}\rightarrow\mathbb{R}^{m},
\qquad
E_y:\mathbb{R}\rightarrow\mathbb{R}^{m},
\end{equation*}
where in our particular case $n=10$ and $m=512$. For a labeled training row $(x_i,y_i)$, the initial representation is
formed by adding the two embeddings,
\[
z_i^{(0)} = E_x(x_i) + E_y(y_i),
\]
whereas an unlabeled test row is represented only by its feature
embedding, $z_i^{(0)}=E_x(x_i)$. Thus, training and test rows share the
same $m$-dimensional token space while labels are revealed only for the
training rows.

The resulting sequence is processed by $K=4$ ordered
Hydra--TabPFN pairs, giving eight layers in total:
\[
\underbrace{
  [\,\mathrm{Hydra}\rightarrow\mathrm{TabPFN}\,]
  \times K
}_{\text{2K layers}}.
\]
Every layer preserves the $m$-dimensional representation, allowing
Hydra and attention layers to be interleaved without additional
projection layers. The Hydra layers use the bidirectional state-space
mixer of \citet{hwang2024hydra}, following its adaptation to tabular
prior-fitted networks by \citet{koch2025state}. Each Transformer layer
contains multi-head self-attention followed by a two-layer feed-forward
network with hidden dimension $2m$. Both sublayers use residual
connections and layer normalization.

Finally, the representations corresponding to the test rows are passed
through a two-layer MLP,
\[
\mathbb{R}^{m}\rightarrow\mathbb{R}^{2m}
\rightarrow\mathbb{R}^{C},
\]
with a GELU activation, where $C$ denotes number of classes. The resulting class logits are normalized with
a softmax to obtain predictive class probabilities.

\subsection{Bidirectional State-Space Mixing} 
Tabular data has no canonical row order, \textit{that is,} permuting the rows of $X_\text{Train}$ should not change the model's predictions. This \textit{permutation invariance} distinguishes language from tabular in-context learning. Standard SSMs process a sequence left-to-right. So, naively substituting an SSM with TabPFN's set-transformer layers would break permutation invariance. Hydra layers resolve this by replacing the standard SSM with a bidirectional state-space mixer that lets every row attend to every other row, in either direction, while still admitting linear-time evaluation.

Let $L$ denotes sequence length and $D$ number of channels. Formally, a causal selective SSM (as used in unidirectional
Mamba) applied to $\mathbf{X}\in\mathbb{R}^{L\times D}$ acts as a matrix mixer $\mathbf{Y} = \mathbf{M}\mathbf{X}$,
where entry $m_{ts}$ of the mixer matrix is
\begin{equation}
m_{ts} = \mathbf{c}_t^\top \Big(\textstyle\prod_{k=s+1}^{t}\mathbf{A}_k\Big)\mathbf{b}_s, \qquad t \geq s,
\end{equation}
with data-dependent $\mathbf{A}_k \in \mathbb{R}^{N\times N}$ and $\mathbf{b}_k, \mathbf{c}_k \in \mathbb{R}^N$ the
discretized state, input, and output matrices at position $k$, and $N$ the SSM state dimension. Such $\mathbf{M}$
are $N$-\emph{semiseparable}: every submatrix taken from the lower triangle has rank at most $N$, which is exactly
what permits an $O(L)$ recurrent evaluation, but also forces $\mathbf{M}$ to be strictly causal (zero above the diagonal).

Hydra \citep{hwang2024hydra} lifts this restriction by parameterizing the mixer as a \emph{quasiseparable} matrix,
\begin{equation}
m_{ts} =
\begin{cases}
\mathbf{c}_t^\top \big(\textstyle\prod_{k=s+1}^{t}\mathbf{A}_k\big)\mathbf{b}_s, & t > s, \\[4pt]
\delta_t, & t = s, \\[4pt]
\overleftarrow{\mathbf{c}}_t^\top \big(\textstyle\prod_{k=t}^{s-1}\overleftarrow{\mathbf{A}}_k\big)\overleftarrow{\mathbf{b}}_s, & t < s,
\end{cases}
\end{equation}
i.e., a lower-triangular semiseparable block identical to a causal SSM, an independently parameterized
upper-triangular semiseparable block built from a second, reversed set of state matrices
$\overleftarrow{\mathbf{A}}_k, \overleftarrow{\mathbf{b}}_k, \overleftarrow{\mathbf{c}}_k$, and free diagonal terms
$\delta_t$. Crucially, the rank bound defining quasiseparable matrices holds only for the strictly upper- and
lower-triangular submatrices, not across the diagonal as for semiseparable matrices -- so this parameterization is
strictly more expressive than naive bidirectional SSMs that tie the forward and backward passes together through a
shared diagonal, while still admitting sub-quadratic evaluation.

In practice, the quasiseparable mixer is never materialized directly; instead it is realized by combining two
ordinary causal SSM (semiseparable) passes, $\mathrm{QS}(\mathbf{X}) =$
\begin{equation}
\mathrm{shift}\big(\mathrm{SS}(\mathbf{X})\big)
+ \mathrm{flip}\Big(\mathrm{shift}\big(\mathrm{SS}(\mathrm{flip}(\mathbf{X}))\big)\Big)
+ \mathbf{D}\mathbf{X},
\end{equation}
where $\mathrm{SS}(\cdot)$ denotes a standard causal SSM scan, $\mathrm{flip}(\cdot)$
reverses the input sequence, $\mathrm{shift}(\cdot)$ shifts it one position forward with zero padding at the start,
and $\mathbf{D} = \mathrm{diag}(\delta_1,\dots,\delta_L)$ collects the learned diagonal terms. Any semiseparable SSM
can play the role of $\mathrm{SS}(\cdot)$; following \citet{hwang2024hydra}, we instantiate it with SSD
(structured state-space duality, \citealt{dao2024transformers}), chosen for its linear-time, hardware-efficient scan.
Concretely, each Hydra layer in Tydra runs a forward SSD scan over $\mathbf{X}$ and a second SSD scan over
$\mathrm{flip}(\mathbf{X})$, then re-aligns the reversed output via $\mathrm{flip}\circ\mathrm{shift}$ before summing
both directional contributions with the diagonal (residual) term.
This lets every row attend to every other row in the table, matching the expressivity of full self-attention while retaining $O(L)$ time and memory in the number of rows $L$.

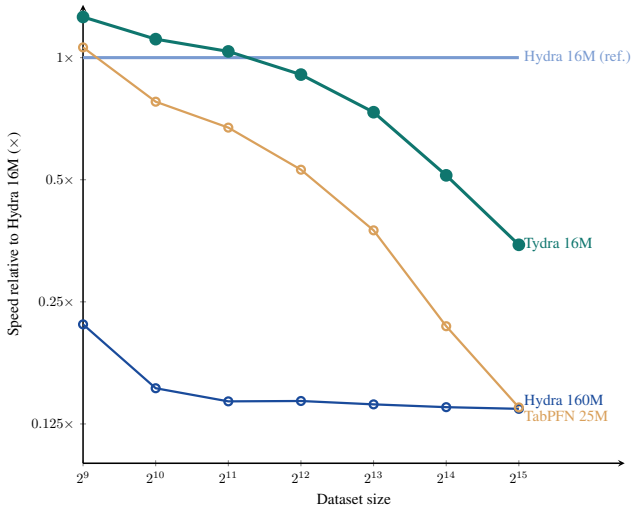
\begin{figure}[t]
\centering
\resizebox{\linewidth}{!}{\begin{tikzpicture}
\begin{axis}[
    xlabel={Dataset size},
    ylabel={Speed relative to Hydra 16M ($\times$)},
    xlabel style={font=\large},
    ylabel style={font=\large},
    xmode=log, log basis x=2,
    ymode=log, log basis y=2,
    xmax=90000,                    
    ymin=0.1, ymax=1.35,
    width=0.85\textwidth,
    xtick={512,1024,2048,4096,8192,16384,32768},
    ytick={0.125,0.25,0.5,1},
    yticklabels={$0.125\times$,$0.25\times$,$0.5\times$,$1\times$},
    tick label style={font=\normalsize},
    axis lines=left,
    line width=1.1pt,              
    clip=false,
]
\draw[color=hydra22Mcolor, line width=2.2pt] (axis cs:512,1) -- (axis cs:32768,1);
\addplot[color=tydracolor, mark=*, mark size=3.4pt, line width=2.2pt, forget plot] coordinates {
(512,1.2597) (1024,1.1107) (2048,1.0359) (4096,0.9082)
(8192,0.7337) (16384,0.5126) (32768,0.3456)
};
\addplot[color=hydra160Mcolor, mark=o, mark size=2.8pt, line width=1.6pt, forget plot] coordinates {
(512,0.2200) (1024,0.1531) (2048,0.1421) (4096,0.1424)
(8192,0.1397) (16384,0.1375) (32768,0.1362)
};
\addplot[color=tabpfncolor, mark=o, mark size=2.8pt, line width=1.6pt, forget plot] coordinates {
(512,1.0599) (1024,0.7788) (2048,0.6722) (4096,0.5293)
(8192,0.3752) (16384,0.2177) (32768,0.1373)
};
\node[font=\large, text=hydra22Mcolor, anchor=west] at (axis cs:32768,1) {Hydra 16M (ref.)};
\node[font=\large, text=tydracolor, anchor=west] at (axis cs:32768,0.3456) {Tydra 16M};
\node[font=\large, text=hydra160Mcolor, anchor=west, yshift=5pt] at (axis cs:32768,0.1362) {Hydra 160M};
\node[font=\large, text=tabpfncolor, anchor=west, yshift=-6pt] at (axis cs:32768,0.1373) {TabPFN 25M};
\end{axis}
\end{tikzpicture}}
\caption{{\bf Tydra scales far better than TabPFN and Hydra 160M, staying close to Hydra 16M's speed well into the large-context regime.} Inference speed of Tydra, Hydra 160M, and TabPFN relative to Hydra 16M on the synthetic benchmark. Tydra is competitive with or faster than Hydra 16M up to $2^{12}$ samples, then falls increasingly behind at larger scales, reaching roughly a third of Hydra 16M's speed at $2^{15}$ -- still far ahead of TabPFN and Hydra 160M. Hydra 160M is substantially slower than Hydra 16M across the entire range, while TabPFN degrades from being competitive at small sizes (up to $1.06\times$ faster) to roughly to roughly the same speed as Hydra 160M at $2^{15}$.}
\label{fig:hybrid_time_rel}
\end{figure}

\begin{figure*}[t]
\centering

\resizebox{\linewidth}{!}{\pgfplotsset{compat=1.18}
\usepgfplotslibrary{groupplots}

\definecolor{tydracolor}{RGB}{31,119,180}
\definecolor{hydra16Mcolor}{HTML}{D9A05B}
\definecolor{hydra160Mcolor}{RGB}{255,127,14}

\begin{tikzpicture}
\begin{groupplot}[
    group style={
        group size=1 by 2,
        vertical sep=12pt,
        x descriptions at=edge bottom,
    },
    width=16.5cm,
    height=6.2cm,
    symbolic x coords={
        31, 50, 1494, 1049, 40966, 1050, 23, 1068, 40975, 1462,
        54, 37, 469, 15, 1464, 458, 11, 29, 188, 14,
        22, 18, 40982, 16, 1063, 40994, 1480, 1510, 6332, 23381
    },
    xtick=data,
    xtick pos=left,
    ytick pos=left,
    x tick label style={rotate=45, anchor=east, font=\tiny},
    ymajorgrids=true,
    grid style={dashed, gray!20},
    y tick label style={font=\footnotesize},
    every axis/.append style={ybar=0pt, bar width=1.8pt},
    legend style={
        at={(0.5,1.22)},
        anchor=south,
        legend columns=-1,
        font=\footnotesize,
        /tikz/every even column/.append style={column sep=6pt}
    },
    clip marker paths=true,
]

\nextgroupplot[
    ylabel={Inference speedup vs.\ TabPFN ($\times$)},
    ylabel style={font=\small},
    ymin=0,
    ymax=1.55,
    ytick={0,0.25,0.5,0.75,1,1.25,1.5},
    legend to name=threemodellegend,
    clip=false,
]

\addplot[
    fill=tydracolor!60,
    draw=tydracolor,
    line width=0.2pt
] coordinates {
    (31,1.426) (50,1.423) (1494,1.423) (1049,1.408)
    (40966,1.403) (1050,1.395) (23,1.394) (1068,1.393)
    (40975,1.391) (1462,1.388) (54,1.388) (37,1.379)
    (469,1.379) (15,1.376) (1464,1.376) (458,1.375)
    (11,1.374) (29,1.368) (188,1.366) (14,1.362)
    (22,1.354) (18,1.351) (40982,1.348) (16,1.346)
    (1063,1.338) (40994,1.308) (1480,1.284) (1510,1.244)
    (6332,1.036) (23381,0.887)
};
\addlegendentry{Tydra \{4 HT \} (ours)}

\addplot[
    fill=hydra16Mcolor!70,
    draw=hydra16Mcolor,
    line width=0.2pt
] coordinates {
    (31,1.083) (50,1.092) (1494,1.115) (1049,1.129)
    (40966,0.989) (1050,1.112) (23,1.108) (1068,1.097)
    (40975,1.120) (1462,1.274) (54,1.006) (37,1.053)
    (469,1.151) (15,1.074) (1464,1.190) (458,1.028)
    (11,1.133) (29,1.050) (188,0.980) (14,0.978)
    (22,0.979) (18,1.109) (40982,1.008) (16,0.976)
    (1063,1.120) (40994,1.030) (1480,1.037) (1510,1.203)
    (6332,0.766) (23381,0.654)
};
\addlegendentry{Hydra 16M}

\addplot[
    fill=hydra160Mcolor!70,
    draw=hydra160Mcolor,
    line width=0.2pt
] coordinates {
    (31,0.196) (50,0.195) (1494,0.195) (1049,0.195)
    (40966,0.175) (1050,0.197) (23,0.176) (1068,0.198)
    (40975,0.173) (1462,0.201) (54,0.175) (37,0.197)
    (469,0.171) (15,0.209) (1464,0.201) (458,0.170)
    (11,0.180) (29,0.211) (188,0.171) (14,0.162)
    (22,0.162) (18,0.165) (40982,0.165) (16,0.162)
    (1063,0.217) (40994,0.205) (1480,0.205) (1510,0.211)
    (6332,0.239) (23381,0.274)
};
\addlegendentry{Hydra 160M}

\draw[black!65, dashed, line width=0.6pt]
    (axis cs:31,1) -- (axis cs:23381,1);

\node[
    font=\scriptsize,
    anchor=south east,
    text=black!65
] at (rel axis cs:0.99,0.98) {higher is better};

\nextgroupplot[
    xlabel={Dataset ID, sorted by Tydra \{4HT\} speedup},
    ylabel={Score change vs.\ TabPFN (pts)},
    ylabel style={font=\small},
    ymin=-16,
    ymax=4,
    clip=false,
]

\addplot[
    fill=tydracolor!60,
    draw=tydracolor,
    line width=0.2pt
] coordinates {
    (31,-0.03) (50,-0.49) (1494,-0.29) (1049,-0.34)
    (40966,0.00) (1050,0.36) (23,-0.57) (1068,0.37)
    (40975,-0.28) (1462,0.00) (54,-0.78) (37,-1.31)
    (469,1.23) (15,-0.11) (1464,-1.12) (458,0.00)
    (11,-0.31) (29,-0.50) (188,-0.33) (14,-0.10)
    (22,-0.18) (18,0.08) (40982,-0.44) (16,-0.04)
    (1063,-3.74) (40994,0.23) (1480,-0.96) (1510,-0.02)
    (6332,1.07) (23381,-0.16)
};

\addplot[
    fill=hydra16Mcolor!70,
    draw=hydra16Mcolor,
    line width=0.2pt
] coordinates {
    (31,-0.57) (50,-14.93) (1494,-1.58) (1049,-1.69)
    (40966,-0.90) (1050,-0.97) (23,-1.23) (1068,-5.22)
    (40975,-4.84) (1462,0.00) (54,-3.29) (37,0.25)
    (469,-0.36) (15,0.01) (1464,0.69) (458,-0.01)
    (11,-1.06) (29,-0.29) (188,-2.30) (14,-1.82)
    (22,-2.16) (18,-0.50) (40982,-3.84) (16,-1.09)
    (1063,-0.27) (40994,-0.54) (1480,-1.46) (1510,0.01)
    (6332,-3.32) (23381,-1.14)
};

\addplot[
    fill=hydra160Mcolor!70,
    draw=hydra160Mcolor,
    line width=0.2pt
] coordinates {
    (31,-0.55) (50,-9.77) (1494,-1.26) (1049,-0.77)
    (40966,-0.89) (1050,-0.51) (23,0.08) (1068,-3.20)
    (40975,-1.34) (1462,-0.00) (54,-2.86) (37,0.22)
    (469,0.27) (15,0.06) (1464,0.67) (458,-0.01)
    (11,-0.34) (29,-0.29) (188,-0.95) (14,-2.01)
    (22,-1.23) (18,-0.46) (40982,-2.85) (16,-1.15)
    (1063,-0.51) (40994,-0.05) (1480,-0.59) (1510,-0.00)
    (6332,-2.81) (23381,0.43)
};

\draw[black!65, dashed, line width=0.6pt]
    (axis cs:31,0) -- (axis cs:23381,0);

\end{groupplot}\node at (group c1r1.north) [anchor=south, yshift=8pt] {\ref{threemodellegend}};

\end{tikzpicture}}

\caption{
\textbf{Tydra matches TabPFN's accuracy at substantially lower inference cost.} 
Inference speedup relative to TabPFN (top) and change in predictive performance ($\Delta
\text{AUROC}$) relative to TabPFN (bottom), for Tydra {4HT} and two Hydra baselines (16M, 160M), across 30 OpenML datasets. Datasets are sorted by Tydra's inference speedup.}\label{fig:openml_barplot}
\end{figure*}
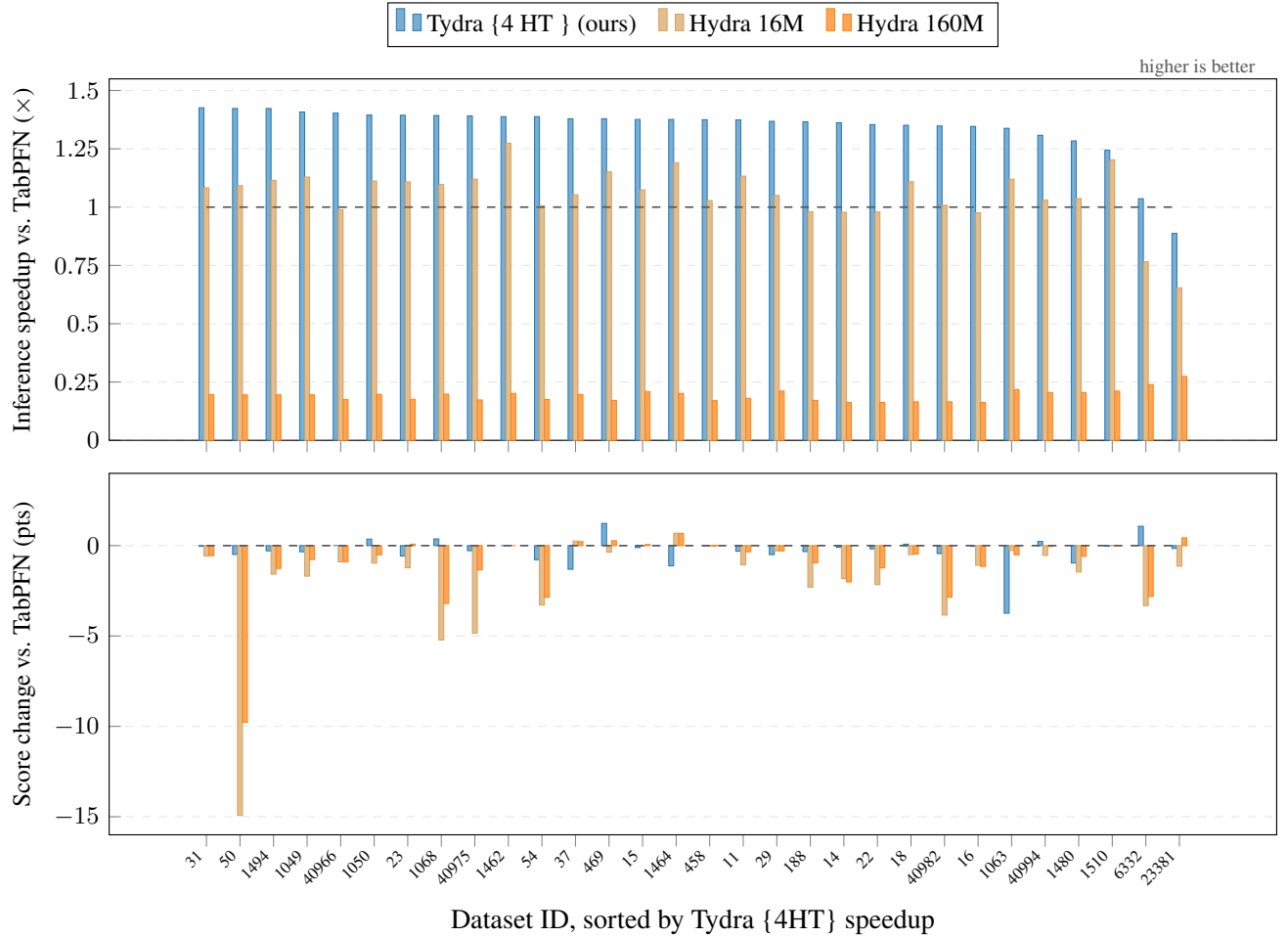

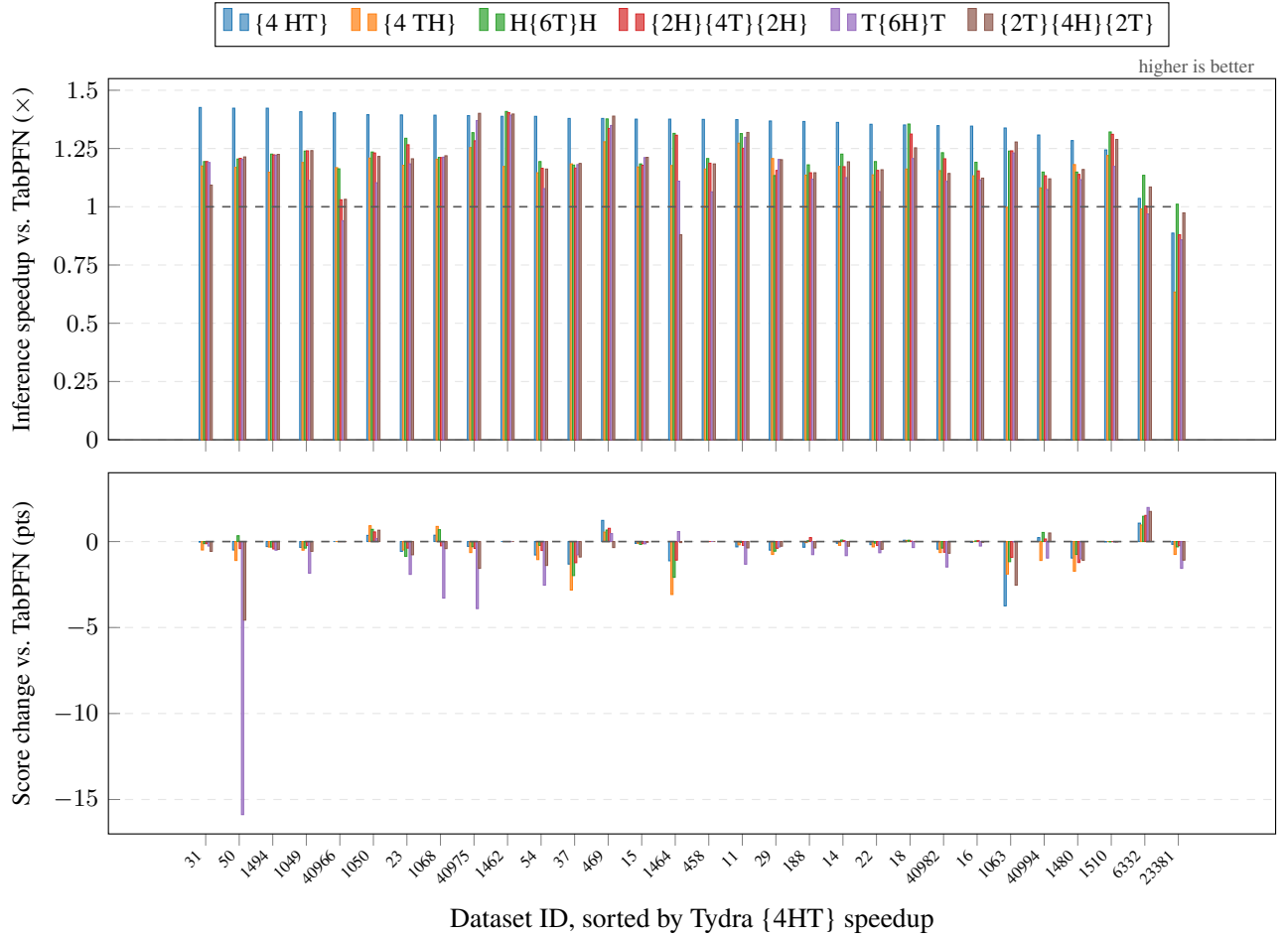
\begin{figure*}[t]
\centering
\resizebox{\linewidth}{!}{\pgfplotsset{compat=1.18}
\usepgfplotslibrary{groupplots}

\definecolor{c8l}{RGB}{31,119,180}
\definecolor{cTH4}{RGB}{255,127,14}
\definecolor{cHTTTTTTH}{RGB}{44,160,44}
\definecolor{cHHTTTTHH}{RGB}{214,39,40}
\definecolor{cTHHHHHHT}{RGB}{148,103,189}
\definecolor{cTTHHHHTT}{RGB}{140,86,75}

\begin{tikzpicture}
\begin{groupplot}[
    group style={
        group size=1 by 2,
        vertical sep=12pt,
        x descriptions at=edge bottom,
    },
    width=16.5cm,
    height=6.2cm,
    symbolic x coords={
        31, 50, 1494, 1049, 40966, 1050, 23, 1068, 40975, 1462,
        54, 37, 469, 15, 1464, 458, 11, 29, 188, 14,
        22, 18, 40982, 16, 1063, 40994, 1480, 1510, 6332, 23381
    },
    xtick=data,
    xtick pos=left,
    ytick pos=left,
    x tick label style={rotate=45, anchor=east, font=\tiny},
    ymajorgrids=true,
    grid style={dashed, gray!20},
    y tick label style={font=\footnotesize},
    every axis/.append style={ybar=0pt, bar width=0.8pt},
    legend style={
        at={(0.5,1.22)},
        anchor=south,
        legend columns=-1,
        font=\footnotesize,
        /tikz/every even column/.append style={column sep=6pt}
    },
    clip marker paths=true,
]

\nextgroupplot[
    ylabel={Inference speedup vs.\ TabPFN ($\times$)},
    ylabel style={font=\small},
    ymin=0,
    ymax=1.55,
    ytick={0,0.25,0.5,0.75,1,1.25,1.5},
    legend to name=hybridlegend,
    clip=false,
]

\addplot[fill=c8l!60, draw=c8l, line width=0.15pt] coordinates {
    (31,1.426) (50,1.423) (1494,1.423) (1049,1.408) (40966,1.403) (1050,1.395) (23,1.394) (1068,1.393) (40975,1.391) (1462,1.388) (54,1.388) (37,1.379) (469,1.379) (15,1.376) (1464,1.376) (458,1.375) (11,1.374) (29,1.368) (188,1.366) (14,1.362) (22,1.354) (18,1.351) (40982,1.348) (16,1.346) (1063,1.338) (40994,1.308) (1480,1.284) (1510,1.244) (6332,1.036) (23381,0.887)
};
\addlegendentry{\{4 HT\}}

\addplot[fill=cTH4!70, draw=cTH4, line width=0.15pt] coordinates {
    (31,1.175) (50,1.169) (1494,1.148) (1049,1.19) (40966,1.168) (1050,1.209) (23,1.178) (1068,1.204) (40975,1.254) (1462,1.173) (54,1.146) (37,1.184) (469,1.279) (15,1.171) (1464,1.177) (458,1.162) (11,1.273) (29,1.207) (188,1.136) (14,1.173) (22,1.137) (18,1.162) (40982,1.155) (16,1.133) (1063,0.998) (40994,1.081) (1480,1.181) (1510,1.22) (6332,0.991) (23381,0.634)
};
\addlegendentry{ \{4 TH\}}

\addplot[fill=cHTTTTTTH!70, draw=cHTTTTTTH, line width=0.15pt] coordinates {
    (31,1.194) (50,1.205) (1494,1.226) (1049,1.239) (40966,1.163) (1050,1.235) (23,1.294) (1068,1.212) (40975,1.318) (1462,1.409) (54,1.194) (37,1.179) (469,1.377) (15,1.184) (1464,1.315) (458,1.207) (11,1.314) (29,1.134) (188,1.18) (14,1.226) (22,1.194) (18,1.355) (40982,1.232) (16,1.191) (1063,1.239) (40994,1.149) (1480,1.149) (1510,1.321) (6332,1.135) (23381,1.011)
};
\addlegendentry{H\{6T\}H}

\addplot[fill=cHHTTTTHH!70, draw=cHHTTTTHH, line width=0.15pt] coordinates {
    (31,1.194) (50,1.208) (1494,1.223) (1049,1.24) (40966,1.03) (1050,1.231) (23,1.266) (1068,1.211) (40975,1.283) (1462,1.404) (54,1.165) (37,1.166) (469,1.336) (15,1.178) (1464,1.307) (458,1.187) (11,1.251) (29,1.156) (188,1.146) (14,1.172) (22,1.156) (18,1.312) (40982,1.206) (16,1.154) (1063,1.24) (40994,1.133) (1480,1.139) (1510,1.311) (6332,1.003) (23381,0.88)
};
\addlegendentry{\{2H\}\{4T\}\{2H\}}

\addplot[fill=cTHHHHHHT!70, draw=cTHHHHHHT, line width=0.15pt] coordinates {
    (31,1.19) (50,1.203) (1494,1.221) (1049,1.113) (40966,0.94) (1050,1.103) (23,1.184) (1068,1.213) (40975,1.37) (1462,1.393) (54,1.078) (37,1.181) (469,1.349) (15,1.211) (1464,1.11) (458,1.064) (11,1.298) (29,1.203) (188,1.118) (14,1.125) (22,1.065) (18,1.207) (40982,1.109) (16,1.115) (1063,1.231) (40994,1.074) (1480,1.116) (1510,1.173) (6332,0.969) (23381,0.858)
};
\addlegendentry{T\{6H\}T}

\addplot[fill=cTTHHHHTT!70, draw=cTTHHHHTT, line width=0.15pt] coordinates {
    (31,1.093) (50,1.214) (1494,1.225) (1049,1.241) (40966,1.032) (1050,1.216) (23,1.206) (1068,1.219) (40975,1.401) (1462,1.398) (54,1.162) (37,1.186) (469,1.389) (15,1.212) (1464,0.88) (458,1.184) (11,1.319) (29,1.202) (188,1.146) (14,1.193) (22,1.159) (18,1.253) (40982,1.143) (16,1.123) (1063,1.278) (40994,1.12) (1480,1.16) (1510,1.289) (6332,1.085) (23381,0.974)
};
\addlegendentry{\{2T\}\{4H\}\{2T\}}

\draw[black!65, dashed, line width=0.6pt]
    (axis cs:31,1) -- (axis cs:23381,1);

\node[
    font=\scriptsize,
    anchor=south east,
    text=black!65
] at (rel axis cs:0.99,0.98) {higher is better};

\nextgroupplot[
    xlabel={Dataset ID, sorted by Tydra \{4HT\} speedup},
    ylabel={Score change vs.\ TabPFN (pts)},
    ylabel style={font=\small},
    ymin=-17,
    ymax=4,
    clip=false,
]

\addplot[fill=c8l!60, draw=c8l, line width=0.15pt] coordinates {
    (31,-0.03) (50,-0.49) (1494,-0.29) (1049,-0.34) (40966,0.0) (1050,0.36) (23,-0.57) (1068,0.37) (40975,-0.28) (1462,0.0) (54,-0.78) (37,-1.31) (469,1.23) (15,-0.11) (1464,-1.12) (458,0.0) (11,-0.31) (29,-0.5) (188,-0.33) (14,-0.1) (22,-0.18) (18,0.08) (40982,-0.44) (16,-0.04) (1063,-3.74) (40994,0.23) (1480,-0.96) (1510,-0.02) (6332,1.07) (23381,-0.16)
};

\addplot[fill=cTH4!70, draw=cTH4, line width=0.15pt] coordinates {
    (31,-0.49) (50,-1.1) (1494,-0.32) (1049,-0.5) (40966,-0.0) (1050,0.92) (23,-0.49) (1068,0.88) (40975,-0.64) (1462,0.0) (54,-1.05) (37,-2.82) (469,0.55) (15,-0.1) (1464,-3.08) (458,-0.0) (11,-0.19) (29,-0.75) (188,-0.06) (14,-0.22) (22,-0.31) (18,0.06) (40982,-0.64) (16,0.0) (1063,-1.89) (40994,-1.1) (1480,-1.73) (1510,-0.0) (6332,0.96) (23381,-0.75)
};

\addplot[fill=cHTTTTTTH!70, draw=cHTTTTTTH, line width=0.15pt] coordinates {
    (31,-0.12) (50,0.34) (1494,-0.32) (1049,-0.38) (40966,0.0) (1050,0.71) (23,-0.86) (1068,0.69) (40975,-0.3) (1462,0.0) (54,-0.22) (37,-1.98) (469,0.67) (15,-0.17) (1464,-2.08) (458,0.0) (11,-0.05) (29,-0.56) (188,0.05) (14,0.09) (22,-0.09) (18,0.08) (40982,-0.38) (16,0.04) (1063,-1.17) (40994,0.53) (1480,-0.75) (1510,-0.02) (6332,1.47) (23381,-0.31)
};

\addplot[fill=cHHTTTTHH!70, draw=cHHTTTTHH, line width=0.15pt] coordinates {
    (31,-0.12) (50,-0.41) (1494,-0.43) (1049,-0.23) (40966,0.0) (1050,0.56) (23,-0.38) (1068,-0.25) (40975,-0.4) (1462,0.0) (54,-0.52) (37,-1.23) (469,0.77) (15,-0.12) (1464,-1.08) (458,0.0) (11,-0.22) (29,-0.39) (188,0.23) (14,0.05) (22,-0.23) (18,0.03) (40982,-0.62) (16,0.05) (1063,-0.92) (40994,0.15) (1480,-1.22) (1510,-0.0) (6332,1.52) (23381,-0.24)
};

\addplot[fill=cTHHHHHHT!70, draw=cTHHHHHHT, line width=0.15pt] coordinates {
    (31,-0.26) (50,-15.88) (1494,-0.5) (1049,-1.84) (40966,-0.0) (1050,0.17) (23,-1.91) (1068,-3.29) (40975,-3.9) (1462,-0.0) (54,-2.54) (37,-0.77) (469,0.46) (15,-0.13) (1464,0.58) (458,-0.0) (11,-1.32) (29,-0.32) (188,-0.76) (14,-0.82) (22,-0.65) (18,-0.35) (40982,-1.48) (16,-0.26) (1063,-0.02) (40994,-0.96) (1480,-1.0) (1510,-0.02) (6332,1.98) (23381,-1.56)
};

\addplot[fill=cTTHHHHTT!70, draw=cTTHHHHTT, line width=0.15pt] coordinates {
    (31,-0.57) (50,-4.57) (1494,-0.46) (1049,-0.57) (40966,0.0) (1050,0.66) (23,-0.77) (1068,-0.41) (40975,-1.56) (1462,0.0) (54,-1.39) (37,-0.9) (469,-0.35) (15,-0.04) (1464,-0.06) (458,-0.0) (11,-0.37) (29,-0.27) (188,-0.37) (14,-0.27) (22,-0.45) (18,0.01) (40982,-0.68) (16,-0.04) (1063,-2.54) (40994,0.5) (1480,-1.09) (1510,-0.02) (6332,1.75) (23381,-1.08)
};

\draw[black!65, dashed, line width=0.6pt]
    (axis cs:31,0) -- (axis cs:23381,0);

\end{groupplot}

\node at (group c1r1.north) [anchor=south, yshift=8pt] {\ref{hybridlegend}};

\end{tikzpicture}}
\caption{\textbf{The Tydra family consistently speeds up inference over TabPFN, with balanced hybrid variants also preserving predictive performance.} Inference speedup (top) and AUC-ROC change (bottom) relative to TabPFN across 30 OpenML datasets, for six Tydra configurations that vary the ratio and placement of attention (T) and Hydra (H) layers. All 
achieve similar inference speedups over TabPFN (up to 
1.4$\times$), but the one that concentrate Hydra layers together, such as T\{6H\}T, show substantially larger accuracy drops on individual datasets (e.g., -15.7 points on dataset 50) than more balanced interleavings. Datasets are sorted by Tydra \{4HT\}'s inference speedup.}
\label{fig:tydras_barplots}
\end{figure*}

\subsection{Training}


We meta-train Tydra offline using the prior-data fitted network
pipeline adopted by \citet{koch2025state}. Training is performed on
synthetic tasks generated on the fly using the neural-network component
of the TabPFN prior framework. A synthetic task
\begin{equation}
    \mathcal{T}_t
    =
    \left(\mathbf{X}_t,\mathbf{y}_t\right)
\end{equation}
is one complete, randomly generated tabular classification dataset,
where $\mathbf{X}_t\in\mathbb{R}^{n_t\times d_t}$ contains its feature
values and $\mathbf{y}_t\in\{1,\ldots,K_t\}^{n_t}$ contains its class
labels.

Let $\boldsymbol{\phi}_t$ collect the hyperparameters of the
neural-network generator used to produce task $t$, including its
architecture, activation functions, dropout probability,
weight-initialization scale, and noise level. We use
$p_{\mathrm{HP}}(\boldsymbol{\phi})$ to denote the configured
hyperprior from which these generator hyperparameters are sampled.
Conditional on $\boldsymbol{\phi}_t$, the distribution
$p_{\mathrm{NN}}(\mathcal{T}\mid\boldsymbol{\phi}_t)$ describes the
datasets produced by the resulting neural-network generator. Task
generation can therefore be written as
\begin{equation}
    \boldsymbol{\phi}_t
    \sim p_{\mathrm{HP}}(\boldsymbol{\phi}),
    \qquad
    \mathcal{T}_t
    \sim
    p_{\mathrm{NN}}
    \left(\mathcal{T}\mid\boldsymbol{\phi}_t\right).
    \label{eq:synthetic_task_generation}
\end{equation}
For each sampled task $\mathcal{T}_t$, the observations are divided into
a labeled context set
\begin{equation}
\mathcal{C}_t = \{(\mathbf{x}_{t,j},y_{t,j})\}_{j=1}^{N_t}
\end{equation}
and a query set
\begin{equation}
\mathcal{Q}_t=\{(\mathbf{x}_{t,i},y_{t,i})\}_{i=1}^{M_t}.
\end{equation}
Given the context and query features, Tydra produces query logits
\begin{equation}
    \mathbf{z}_{t,i}
    =
    f_{\theta}(\mathcal{C}_t,\mathbf{x}_{t,i}).
\end{equation}
For a minibatch of $B$ tasks, we minimize the unweighted multiclass
cross-entropy $ \mathcal{L}(\theta)=$
\begin{equation}
    -\frac{1}{B} \sum_{t=1}^{B}
    \frac{1}{M_t}
    \sum_{i=1}^{M_t}
    \log
    \left[
        \operatorname{softmax}(\mathbf{z}_{t,i})
    \right]_{y_{t,i}} .
    \label{eq:meta_training_objective}
\end{equation}
Thus, the loss is averaged over the query rows of each task and over
the tasks in the minibatch.

At inference time, the
labeled training data form the context $\mathcal{C}$, while the test
features form the queries. Using the fixed meta-trained parameters
$\theta^{\star}$, the predictive distribution for a test example is
\begin{equation}
    p_{\theta^{\star}}
    (y_i=k \mid \mathbf{x}_i,\mathcal{C})
    =
    \left[
        \operatorname{softmax}
        \left(
            f_{\theta^{\star}}(\mathcal{C},\mathbf{x}_i)
        \right)
    \right]_k ,
\end{equation}
and the predicted class is
\begin{equation}
    \hat{y}_i
    =
    \arg\max_k
    p_{\theta^{\star}}
    (y_i=k \mid \mathbf{x}_i,\mathcal{C}).
\end{equation}


We largely follow the training protocol of \citet{koch2025state}, retaining its synthetic-task generation and classification objective as well as AdamW optimization with a learning rate of \(10^{-4}\), an effective batch size of 64, and gradient-norm clipping at \(1.0\), while replacing the original Transformer backbone with Tydra's alternating Hydra--Transformer
backbone.
We additionally tighten the synthetic MLP-prior distributions by reducing the upper bound on the sampled mean of the weight-initialization scale from \(10.0\) to \(1.5\) and that of the noise standard deviation from \(0.3\) to \(0.1\). We clamp synthetic-prior activation values to \([-10^4,10^4]\).



\section{Empirical Evaluation}

We aim to investigate the efficiency of tabular hybrid models empirically. To this end, we conduct experiments to tackle the following questions:

\begin{description}
    \item[(Q1)] Does Tydra match TabPFN's predictive performance at lower inference cost? 
   

    \item[(Q2)] Does naively scaling the pure SSM Hydra architectures narrow the accuracy–efficiency gap?
    
    
    \item[(Q3)] How does Tydra's performance vary across different Transformer–SSM interleaving ratios?
\end{description}


\subsection{Experimental Setup} 
\textbf{Datasets.}
We evaluated Tydra in two regimes. First, we evaluated it on the 30 binary and multiclass
classification datasets from OpenML-CC-18
\citep{bischl2017openml} containing at most 2,000 observations, 100
features, and 10 classes, following
\citet{koch2025state}.
Second, we evaluated its inference speed at longer context lengths using synthetic
tabular classification datasets with table sizes ranging from 512 to 32,768
samples. Unless stated otherwise, each dataset contains 10 numerical features
drawn independently from a standard normal distribution and two balanced
classes assigned independently of the features. Samples are jointly shuffled
using a fixed seed to ensure reproducibility. Each table is divided equally
into a context set and a query set: a table of size $N$ contains $N/2$ context
samples and $N/2$ samples for which predictions are produced. The synthetic
datasets contain no categorical features or missing values and are used solely
to measure inference speed and memory scaling.


\textbf{Methods.} We compare Tydra with pure Transformer and pure SSM baselines. As a Transformer baseline, we use the 25.8M-parameter TabPFN-style architecture following \citet{koch2025state}. As an SSM baseline, we use Hydra at two scales: 16M and 160M parameters. Finally, we evaluate a family of Tydra hybrid models that vary the ratio and placement of attention and Hydra layers ( Fig~\ref{fig:tydra_architectures}). All models are trained using the same prior-fitting procedure and synthetic-data prior introduced by~\citet{koch2025state}.

\textbf{Metrics.}
We evaluated all models in terms of predictive performance and inference
speed. We measured predictive performance using the area under the receiver
operating characteristic curve (AUROC). For each dataset, we evaluated five
deterministic 50/50 train-test splits in which both partitions contain all
observed classes. We report the mean AUROC in the main results and the
corresponding standard deviation across splits in the appendix.

We define inference speed as the total number of test predictions divided by
the total synchronized model inference time. Following two warm-up runs, we
measured inference time over 75 trials, consisting of 15 repetitions for each
of the five splits. 

For the long-context experiment, we measured synchronized end-to-end
evaluation latency. For each model and table size, we performed two warm-up runs followed by 10 timed repetitions and report the mean latency in seconds.
These measurements included data preparation, model execution,
postprocessing, and metric computation, but excluded model loading and
synthetic dataset generation.

The corresponding standard deviations are
reported in the appendix. Together, the two evaluation regimes characterize
model throughput on real-world datasets and end-to-end computational scaling
as the context length increases.

\subsection{(Answer Q1)} To evaluate whether Tydra matches TabPFN's predictive performance at lower inference cost, we compared Tydra against TabPFN across 30 datasets from OpenML benchmark and a synthetic benchmark isolating runtime scaling. Fig.~\ref{fig:quadrants} presents the accuracy–speed trade-off the OpenML datasets, showing Tydra achieves inference speedups of up to 29.9\% (mean +24.8\%) while matching TabPFN's predictive performance, with differences remaining within a narrow band (mean $|\Delta\text{AUROC}| = 0.006$) on all but one dataset. Fig.~\ref{fig:openml_barplot} breaks these results down per dataset, confirming the speedup is consistent across nearly the full suite rather than driven by a few outliers. For example, dataset 23381 is very small, consisting of 98 datapoints in total, which are split into 49 training points and 49 query points. Thus, Tydra matches TabPFN's predictive performance while substantially reducing inference cost. 



\subsection{(Answer Q2)} To evaluate whether naively scaling pure SSM architectures closes the accuracy–efficiency gap, we compare three Hydra configurations (16M and 160M parameters) with TabPFN and Tydra, and isolate runtime scaling on a synthetic benchmark of up to $2^{15}$ rows. Fig.~\ref{fig:quadrants} positions all four models on accuracy and inference speed jointly, showing Hydra 16M is fast but accuracy-poor, while Hydra 160M is slower than TabPFN itself. Figure~\ref{fig:hybrid_time_rel} isolates runtime scaling on the synthetic benchmark, showing that Tydra scales substantially better than TabPFN and Hydra 160M, while remaining competitive with the fastest model, Hydra 16M, up to $2^{13} $ rows before falling behind at the largest scales; Hydra 16M's speed, however, comes at a substantial cost in predictive performance, as shown in Figs.~\ref{fig:quadrants} and \ref{fig:openml_barplot}. No single Hydra scale or size achieves both TabPFN-level accuracy and a meaningful efficiency advantage. Therefore, naively scaling Hydra fails to close the accuracy–efficiency gap.



\subsection{(Answer Q3)} 
To investigate how the Transformer-SSM composition affects the
accuracy--speed trade-off, we evaluate six eight-layer Tydra variants with
different layer ratios and interleaving patterns, as illustrated in
Figure~\ref{fig:tydra_architectures}. 
Table~\ref{tab:hybrid_vs_tabpfn} summarizes the results of the evaluation. Hybridization generally yields models faster than TabPFN without sacrificing too much predictive power. The $\{4\,\mathrm{HT}\}$ architecture combines both types of layers in a 1:1 ratio and provides the best accuracy--speed trade-off. The $T\{6H\}T$ achitecture, with 6 hydra layers and 2 transformer layers, sacrifices the most predictive power on average at 1.24 points worse than TabPFN. However, disaggregating these metrics across the OpenML datasets (Figure~\ref{fig:tydras_barplots}) shows that that the drop in performance is driven by  Dataset 50, on which $T\{6H\}T$ performs 15 AUC-ROC points worse than TabPFN. Overall, this confirms that hybrid architectures are fast without sacrificing much predictive power, with hybrids closer to 1:1 achieving a better tradeoff.
\section{Conclusions and Future Work}

We introduced Tydra, a hybrid Hydra-Transformer architecture for tabular in-context learning that achieves upto 1.43 $\times$ faster inference than TabPFN while retaining comparable predictive performance across OpenML benchmarks. These results suggest that combining subquadratic sequence mixers with PFN-style learners is a promising direction for scaling tabular foundation models. 

Tydra opens up several directions for future work. First, other subquadratic sequence mixers, such as gated linear attention or Gated DeltaNet~\citep{yang2025gated,merrill2026olmo}, could be adapted for tabular data and interleaved with attention layers in place of Hydra, potentially offering a different point on the accuracy--efficiency trade-off. Second, extending Tydra to large-scale, long-context tabular data remains an important direction. Third, looped transformers~\cite{dehghani2018universal,balef2026one}, which repeatedly apply a shared set of layers rather than stacking distinct ones, offer an orthogonal route to efficient inference, and could be combined with hybridization. Together with these directions, the Tydra family of models points towards a broader design space for efficient hybrid tabular foundation models.

\begin{table}[t!bh]
\centering\small
\caption{\textbf{Hybrid Model Comparison vs.\ TabPFN.} Mean test AUC-ROC and mean inference time (per batch, in milliseconds) across all OpenML datasets. 5/6 hybrid variants achieve faster inference without sacrificing much accuracy.}
\resizebox{\linewidth}{!}{\begin{tabular}{lrr}
\toprule
\textbf{Method} & \textbf{AUROC (\%)} & \textbf{Time (ms)} \\
\midrule
TabPFN  & \applygradientgreen{88.90}{}{none}{50}{88.9}\phantom{(-0.29} & \applygradient{27.12}{}{none}{32.40}{19.99}\phantom{(-0.29} \\
\midrule
Tydra \{4 HT\}       & \applygradientgreen{88.61}{\textcolor{purple}{(-0.29)}}{none}{50}{88.9} & \applygradient{19.99}{\textcolor{teal}{(-7.13)}}{none}{32.40}{19.99} \\
Tydra \{4 TH \} & \applygradientgreen{88.41}{\textcolor{purple}{(-0.49)}}{none}{50}{88.9} & \applygradient{23.41}{\textcolor{teal}{(-3.71)}}{none}{32.40}{19.99} \\
Tydra H\{6T\}H   & \applygradientgreen{88.73}{\textcolor{purple}{(-0.17)}}{none}{50}{88.9} & \applygradient{21.98}{\textcolor{teal}{(-5.14)}}{none}{32.40}{19.99} \\
Tydra \{2H\}\{4T\}\{2H\}   & \applygradientgreen{88.71}{\textcolor{purple}{(-0.19)}}{none}{50}{88.9} & \applygradient{22.57}{\textcolor{teal}{(-4.55)}}{none}{32.40}{19.99} \\
Tydra T\{6H\}T       & \applygradientgreen{87.66}{\textcolor{purple}{(-1.24)}}{none}{50}{88.9} & \applygradient{23.59}{\textcolor{teal}{(-3.63)}}{none}{32.40}{19.99} \\
Tydra \{2T\}\{4H\}\{2T\}  & \applygradientgreen{88.37}{\textcolor{purple}{(-0.53)}}{none}{50}{88.9} & \applygradient{22.78}{\textcolor{teal}{(-4.36)}}{none}{32.40}{19.99} 
\end{tabular}}
\label{tab:hybrid_vs_tabpfn}
\end{table}

\section{Acknowledgement}
We gratefully acknowledge support from the BMFTR project XEI (grant number 16IS24079B), the Cluster of Excellence ``Reasonable AI'' funded by the German Research Foundation (DFG) under Germany’s Excellence Strategy (EXC-3057), the DYNAMIC center funded by the LOEWE program of the Hessian Ministry of Science and Arts (grant number LOEWE1/16/519/03/09.001(0009)/98), German Federal Ministry for Economic Affairs and
Energy (BMWE) through EU-SAI: Souveräne KI für Europa (grant number 13IPC040G), and the US Army Research Office (award W911NF2010224).

\bibliography{refs}



\end{document}